# Uncertainty Quantification for AI-Driven Crash Simulation Surrogates: A Comparative Study of Monte Carlo Dropout and Deep Ensembles on an Open-Source Bumper Beam Benchmark

**Sudeep Chavare**

## Abstract

Machine learning surrogate models are increasingly being explored in engineering product development to augment simulation-driven design, offering near-instantaneous predictions that complement computationally expensive high-fidelity analyses. However, a critical gap limits their adoption in safety-critical workflows: a point prediction without an accompanying uncertainty estimate cannot tell an engineer when the model should not be trusted. This work presents a systematic, head-to-head comparison of two widely used uncertainty quantification approaches — Monte Carlo Dropout and Deep Ensembles — applied to an open-source surrogate pipeline built on NVIDIA PhysicsNeMo. A key contribution is the use of concrete dropout, a built-in PhysicsNeMo capability that eliminates the dropout rate as a manual hyperparameter by learning it end-to-end during training, directly addressing the most common criticism of Monte Carlo Dropout-based uncertainty quantification. Automotive crash simulation is used as the application domain, with a steel bumper beam impact problem serving as the benchmark. Both methods are evaluated on identical held-out simulations and compared on point accuracy, uncertainty band calibration, and computational cost. The results reveal a fundamental trade-off between accuracy and calibration that challenges the common assumption that deep ensembles are the default gold standard for surrogate uncertainty quantification. The findings demonstrate that well-calibrated, hyperparameter-free uncertainty estimates are achievable within a fully open-source engineering workflow at a fraction of the computational cost of ensemble approaches.

## 1. Introduction

Crashworthiness assessment is among the most computationally demanding workloads in automotive computer-aided engineering (CAE). A single full-vehicle explicit finite element simulation can require thousands of CPU-hours, and design exploration studies routinely demand hundreds of such runs. Machine learning (ML) surrogate models have emerged as a promising acceleration path, learning the mapping from design parameters to structural response directly from simulation data and providing predictions in milliseconds rather than hours [3]. Recent work has demonstrated that transformer-based neural operators such as Transolver [5] and its geometry-aware extension GeoTransolver [6] can predict full transient deformation fields for industrial crash models with engineering-relevant accuracy [3, 4].

Despite this progress, a critical barrier separates research demonstrations from deployment in safety-critical design workflows: trust. A surrogate model that predicts a peak intrusion of 180 mm is of limited use to a crash engineer unless it can also communicate how confident it is in that number. Decisions about occupant safety cannot rest on point estimates from a black-box model whose failure modes are silent. Uncertainty quantification (UQ), the practice of equipping model predictions with calibrated confidence intervals, is therefore not an academic luxury but a prerequisite for the responsible industrial adoption of AI-driven crash CAE.

The deep learning literature offers two dominant, practical UQ approaches: Monte Carlo (MC) Dropout [8], which interprets dropout regularization as approximate Bayesian inference and extracts uncertainty from stochastic forward passes of a single model, and Deep Ensembles [13], which train multiple independently initialized models and use their disagreement as an uncertainty measure. Deep ensembles are widely regarded as the empirical gold standard but cost as many training runs as ensemble members; MC Dropout costs a single training run but has historically been criticized for its sensitivity to the manually chosen dropout rate. The concrete dropout relaxation [9] addresses this criticism by making per-layer dropout rates learnable parameters optimized jointly with the network weights. Prior work by the author examined ensemble methods and Gaussian Process Regression for uncertainty quantification in machine learning within an automotive engineering context [16], establishing a foundation that the present study extends to physics-informed neural operators for transient crash simulation.

This paper contributes a controlled, head-to-head comparison of concrete MC Dropout and a ten-member deep ensemble on a transient crash surrogate, with three distinguishing features. First, the entire pipeline is built from open-source, zero-cost components: the OpenRadioss explicit FE solver [1] generates the data, the NVIDIA PhysicsNeMo framework [2] provides the GeoTransolver architecture and its native concrete dropout implementation, and Google Colab provides the GPU compute. The study is therefore fully reproducible by any engineer or student without commercial licenses. Second, both UQ methods share an identical architecture, dataset, optimizer, and epoch budget, isolating the uncertainty mechanism as the only experimental variable. Third, the evaluation focuses on the engineering quantity that matters for crash assessment, namely the displacement history of a key performance indicator (KPI) node on the impacted face of the structure, and reports calibration as empirical coverage of the FE ground truth, the metric a practicing engineer would actually rely on.

The remainder of the paper is organized as follows. Section 2 describes the bumper beam benchmark and dataset. Section 3 details the GeoTransolver surrogate architecture. Section 4 presents the two UQ methods with their mathematical foundations. Section 5 describes the experimental setup, and Section 6 reports results. Section 7 discusses the engineering implications of the observed calibration gap, and Section 8 concludes.

## 2. Benchmark Problem and Dataset

The benchmark structure is a steel bumper beam subjected to full-frontal barrier impact at fixed velocity, modeled in OpenRadioss [1], the open-source explicit finite element solver released by Altair Engineering. The model is distributed publicly through openradioss.org and serves as a reproducible community benchmark for structural impact problems. As seen in Figure 1the beam assembly consists of two welded high-strength steel components with dual-phase material grades DP600 and DP1000, discretized with Belytschko-Tsay shell elements. The simulation mesh contains 13,882 nodes, and each transient analysis captures the full impact event, exporting nodal displacement fields at 101 uniformly spaced time states. A single OpenRadioss run requires approximately 15 minutes of wall-clock time on a desktop workstation, establishing the baseline cost that the surrogate model is designed to replace for design-space exploration.

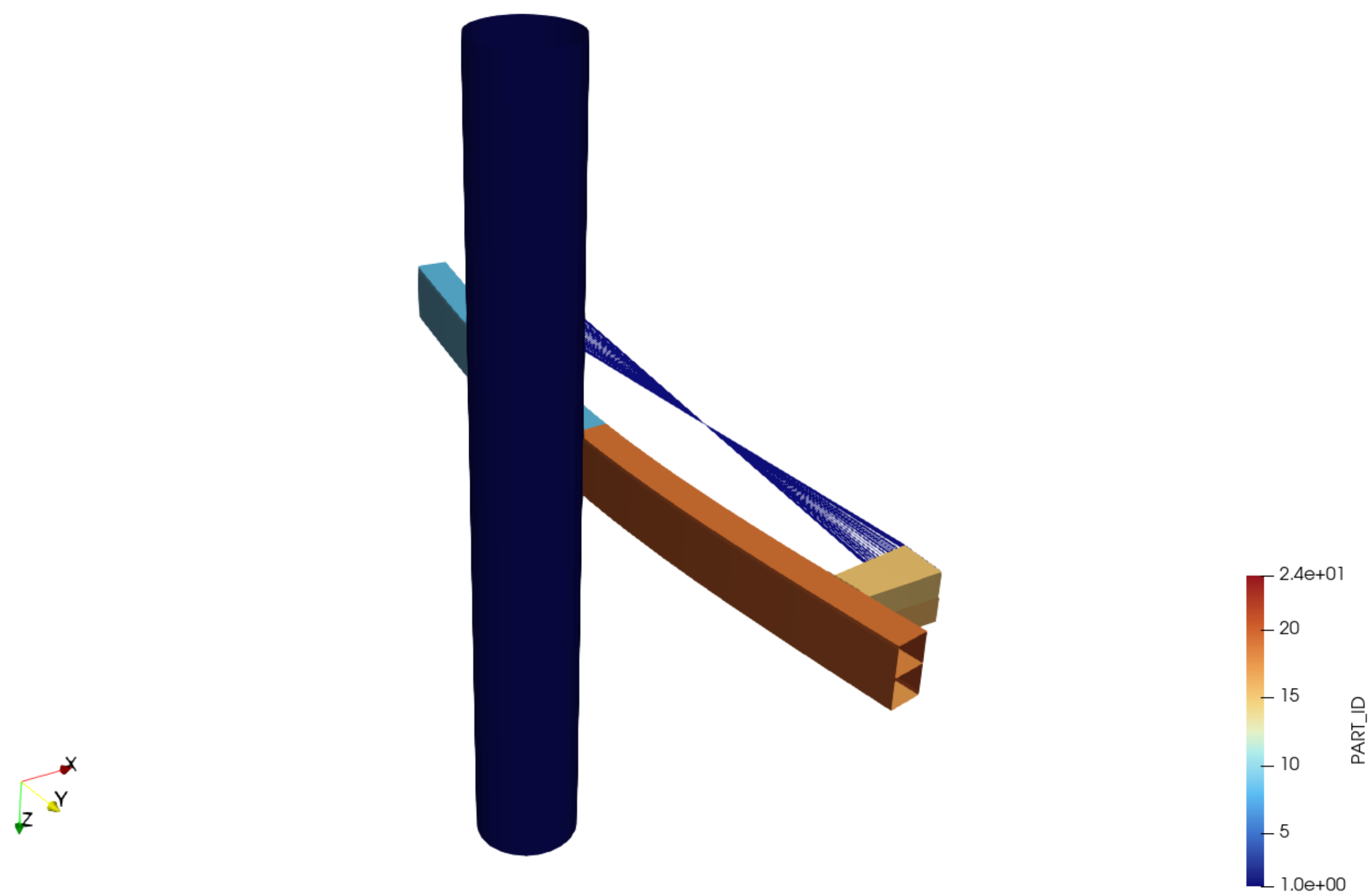

*Figure 1: Public Domain Model of a Bumper Beam crashing a Barrier*

The design space comprises two variables: the shell thickness of the DP600 component (t_DP600) and the shell thickness of the DP1000 component (t_DP1000), with nominal values of 2.0 mm and 2.5 mm respectively. Both thicknesses were varied independently within a ±30% corridor of their nominal values, yielding t_DP600 in [1.4, 2.6] mm and t_DP1000 in [1.75, 3.25] mm. A Latin Hypercube Sampling (LHS) scheme was used to generate 90 design points, providing near-uniform coverage of the two-dimensional parameter space with a minimum number of simulation runs. LHS was preferred over full-factorial or random sampling because it guarantees that no two points share the same value in any single dimension, maximizing the information content of the training set relative to its size. All 90 OpenRadioss simulations were executed and their outputs post-processed using the anim_to_vtk utility, which converts the binary ANIM output format to VTK-readable displacement arrays at each time state. Raw displacement histories were first stored in HDF5 format for exploratory analysis and verification against ParaView visualizations. The data were then converted to VTP (VTK PolyData) files for consumption by the PhysicsNeMo training pipeline. Each VTP file encodes the reference mesh node coordinates as point positions, the per-node shell thickness values as scalar point-data arrays broadcast uniformly to all 13,882 nodes, and the displacement vector field (U_x, U_y, U_z) at each of the 101 time states as individually named point-data arrays. This mesh-native format allows the PhysicsNeMo dataloader to ingest simulation data without index mapping or coordinate interpolation, preserving the full geometric context of the unstructured FEA mesh. Eighty-five runs are used for training and five (Exp_5, Exp_16, Exp_33, Exp_37, Exp_83) are held out as an unseen test set; the identical 85/5 split is used for both UQ methods to ensure a valid comparison.

The engineering KPI tracked throughout this study is the X-direction displacement history of node 1806, located at coordinates (38.93, 0.0, 75.0) mm on the front (impacted) face of the beam. This node exhibits peak X-displacements in the range of approximately -175 to -185 mm across the design space and serves as the primary intrusion metric a crash engineer would monitor. All UQ results reported in this paper are evaluated on the node 1806 displacement time history across the 101 timesteps of each held-out simulation.

### *2.3 Baseline Surrogate: Initial PhysicsNeMo Results Without UQ*

Prior to the UQ study, a baseline deterministic surrogate was trained on the same 85-run dataset using the PhysicsNeMo framework. This initial model used the standard Transolver architecture with a PhysicsNeMo training loop running on a Google Colab T4 GPU over 500 epochs with cosine annealing learning rate schedule. The training employed a dual-objective loss function combining a full-field relative L2 term over all nodes and an auxiliary loss term on node 1806 weighted at LAMBDA_NODE = 2.0, prioritizing accuracy at the engineering KPI location. This KPI-weighted formulation was motivated by the observation that uniform field loss alone underweights the critical front-face intrusion node relative to nodes in low-deformation regions of the beam. Total training wall time was approximately 2.5 hours on the free Colab T4.

On a held-out test case with unseen design parameters (t_DP600 = 2.149 mm, t_DP1000 = 1.456 mm), the baseline surrogate predicted a node 1806 peak X-displacement of approximately -269 mm against an OpenRadioss ground truth of -260 mm, yielding a 3.5% error on peak displacement as seen in Figure 2. The predicted and simulated curves exhibited the same characteristic crash morphology: steep initial deceleration in the first 30 timesteps during barrier contact, a gradual plateau through timesteps 70-90 as the section fully crushes, and peak intrusion at the final timestep. While the curve shape agreement was strong in the early and mid phases, a modest drift was observed in the late timesteps (t = 70-100), where the predicted trajectory deviated slightly from the FEA truth. This late-timestep drift became a primary motivation for the UQ study: if the surrogate systematically underperforms in a specific temporal regime, calibrated uncertainty bands should reflect that, widening in precisely the region where the model is least reliable. A point prediction cannot convey this limitation; UQ does. The inference pipeline from the baseline model was implemented as a Gradio web application with interactive thickness sliders, a timestep scrubber, PyVista surface mesh visualization of the full displacement field, and a GIF export capability for animating the crash sequence. This application confirmed that the surrogate was sufficiently accurate for screening use cases but underscored the absence of any confidence metric alongside the predictions.

The UQ study presented in this paper replaces the baseline Transolver with GeoTransolver and trains from scratch, rather than adding dropout post-hoc to the existing checkpoint. This clean-slate approach ensures that concrete dropout rates are learned from the beginning of training, allowing the regularizer and the data loss to co-optimize throughout the full 200-epoch schedule. The baseline Transolver results serve as context for the surrogate accuracy range achievable on this problem, while the GeoTransolver UQ models reported in Sections 5-6 represent the primary contributions of this work.

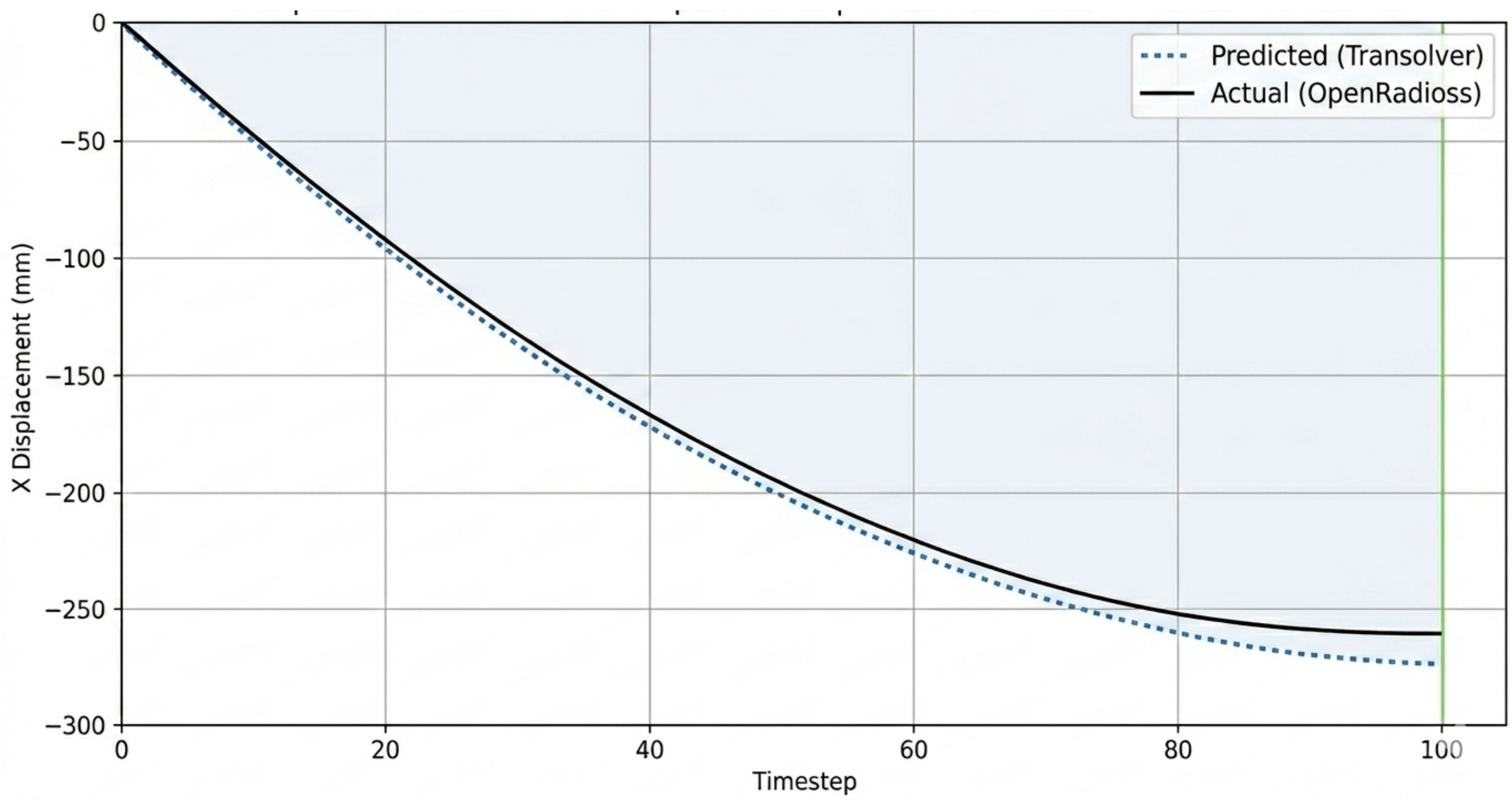

*Figure 2: Comparison of displacement predicted vs actual for node id 1806*

## 3. Surrogate Model Architecture

The surrogate is GeoTransolver [6], a multiscale geometry-aware physics attention transformer released in NVIDIA PhysicsNeMo [2]. GeoTransolver extends the Transolver architecture [5], whose physics-attention mechanism compresses nodal states into a small set of learnable physical slices, performs global interaction in slice space, and maps the result back to the nodes, achieving linear complexity in mesh size. GeoTransolver augments every transformer block with GALE (Geometry-Aware Local Embeddings) cross-attention to a persistent geometry context, anchoring the latent computation to the spatial structure of the domain. Prior work, including studies co-authored by the present author, has demonstrated the effectiveness of this architecture family for transient crash dynamics on both component-level and full-vehicle models [3, 4].

In this study the model operates in a one-shot spatiotemporal configuration: a single forward pass maps the input mesh and design parameters to the complete displacement history. Each node carries a five-dimensional input feature vector (three reference coordinates plus the two component thicknesses), the nodal coordinates serve as the geometry context for GALE cross-attention, and the output is the 303-dimensional flattened displacement trajectory (101 timesteps x 3 components) at every node. Table 1 summarizes the configuration. The identical 6.07M-parameter architecture is used for both UQ methods; the only difference is the presence or absence of concrete dropout layers.

| Parameter | Value |
| --- | --- |
| Architecture | GeoTransolver (PhysicsNeMo v2.0, experimental module) |
| Input features (functional_dim) | 5 (x, y, z, t_DP600, t_DP1000) |
| Geometry context (geometry_dim) | 3 (nodal coordinates) |
| Output dimension | 303 (101 timesteps x 3 displacement components) |
| Transformer blocks | 8 |
| Hidden dimension | 256 |
| Attention heads | 8 |
| Physics-attention slices | 32 |
| Trainable parameters | 6.07 M |
| Multi-scale ball-query local features | Disabled (single-part geometry) |

*Table 1. GeoTransolver configuration used for both UQ studies.*

## 4. Uncertainty Quantification Methods

### *4.1 Monte Carlo Dropout with Learned (Concrete) Rates*

Dropout was introduced by Srivastava et al. [7] as a regularization technique: during training, each unit's activation is zeroed with probability p, preventing co-adaptation of features. Gal and Ghahramani [8] later showed that a network trained with dropout can be interpreted as an approximate Bayesian neural network, where each stochastic forward pass corresponds to a sample from an approximate posterior over the weights. Keeping dropout active at inference and performing T stochastic forward passes yields a predictive distribution. The predictive mean and (epistemic) variance are estimated as:

$$\mu_{MC}(x) = \frac{1}{T}\sum_{t=1}^{T} f(x; W_t) \quad (1)$$

$$\sigma_{MC}{}^2(x) = \frac{1}{T}\sum_{t=1}^{T}(f(x; W_t))^2 - \mu_{MC}(x)^2 \quad (2)$$

where $W_t$ denotes the weights with the t-th sampled dropout mask applied. The principal criticism of MC Dropout is that the quality of the uncertainty estimate depends strongly on the dropout probability p, a hand-tuned hyperparameter: too small and the predictive variance collapses, too large and the model degrades [10]. Concrete dropout [9] resolves this by replacing the discrete Bernoulli mask with a continuous relaxation,

$$z = sigmoid\left(\frac{1}{\tau}[log\, p - log(1-p) + log\, u - log(1-u)]\right), \quad u \sim Uniform(0,1) \quad (3)$$

which is differentiable with respect to p, allowing every dropout layer to learn its own rate by gradient descent. Following the PhysicsNeMo implementation, the training objective augments the data loss with an entropy regularizer over all L dropout layers:

$$\mathcal{P} = \frac{\|\hat{y}-y\|_2}{\|y\|_2} + \lambda\sum_{l=1}^{L}(p_l log p_l + (1-p_l)log(1-p_l)) \quad (4)$$

where the first term is the relative L2 data loss and the regularizer (weighted by lambda = 1e-3) prevents the learned rates from collapsing to trivial solutions. In this study the GeoTransolver model contains 25 concrete dropout layers: three per transformer block (self-attention, GALE cross-attention output, and feed-forward sublayers) across eight blocks, plus one on the geometry tokenizer. All rates are initialized at p = 0.10. At inference, the model is placed in evaluation mode everywhere except the concrete dropout layers, which remain stochastic; T = 50 forward passes per test case produce the predictive distribution of Equations (1)-(2), and the reported uncertainty band is $\mu \pm 2\sigma$.

### *4.2 Deep Ensembles*

Ensembling long predates deep learning: Hansen and Salamon [11] demonstrated in 1990 that committees of neural networks outperform individual members, and Breiman's bagging [12] formalized variance reduction through model averaging. Lakshminarayanan et al. [13] adapted the idea to modern deep networks as a deliberately simple, non-Bayesian UQ method: train M identical architectures from M different random initializations, and treat the spread of their predictions as epistemic uncertainty. Because independently initialized networks descend into different basins of the loss landscape [14], their disagreement reflects genuine ambiguity in what the training data can determine. The ensemble predictive mean and variance are:

$$\mu_{ENS}(x) = \frac{1}{M}\sum_{m=1}^{M} f_m(x) \tag{5}$$

$$\sigma_{ENS}{}^2(x) = \frac{1}{M}\sum_{m=1}^{M}(f_m(x) - \mu_{ENS}(x))^2 \tag{6}$$

Deep ensembles are widely regarded as the strongest practical baseline for deep learning UQ, are hyperparameter-free, and produce deterministic members that can be evaluated in a single pass each. Their cost is the obvious one: M complete training runs. In this study M = 10 deterministic GeoTransolver models (concrete dropout disabled, dropout rate 0.0) were trained with random seeds 0 through 9 controlling both weight initialization and training data shuffling order. The reported band is $\mu \pm 2\sigma$ across the ten members.

### *4.3 Calibration Metric*

Both methods are evaluated by empirical coverage on the KPI node trajectory: the percentage of the 101 timesteps at which the FE ground-truth displacement falls inside the predicted $\mu \pm 2\sigma$ band,

$$C = \frac{100}{N_t}\sum_{t=1}^{N_t} 1\big(y_t \in (\mu_t - 2\sigma_t,\, \mu_t + 2\sigma_t)\big) \tag{7}$$

For a perfectly calibrated Gaussian predictive distribution the expected coverage of a $\pm 2\sigma$ interval is approximately 95%. Coverage far below this target indicates overconfidence (bands too narrow); coverage at or near 100% with reasonable band widths indicates conservative but trustworthy uncertainty estimates [15]. Mean band width is reported alongside coverage, since coverage alone can be trivially achieved with arbitrarily wide bands.

## 5. Experimental Setup

All experiments were conducted on Google Colab using NVIDIA PhysicsNeMo v2.0 (the v2.x release line introduces the experimental GeoTransolver module with native concrete dropout support). The MC Dropout model was trained on a Colab T4 GPU (16 GB) in approximately 2.6 hours; the ten ensemble members were trained sequentially on a Colab Pro A100 GPU (40 GB) at roughly 60 minutes per member. Both studies use identical data normalization (per-feature standardization of inputs; displacement targets scaled by the global maximum of 368 mm), identical optimizer settings, and the identical 85/5 train/test split. Table 2 summarizes the configuration of both training campaigns. Intermediate checkpoints were written to Google Drive every 10 epochs, and the ensemble training loop included automatic skip logic permitting interrupted sessions to resume without retraining completed members.

| Setting | MC Dropout Model | Deep Ensemble (each of 10) |
|---|---|---|
| Concrete dropout | Enabled (25 layers, init p = 0.10) | Disabled |
| Loss | Relative L2 + concrete-dropout regularizer (lambda = 1e-3) | Relative L2 |
| Optimizer | AdamW, lr 1e-4, weight decay 1e-4 | AdamW, lr 1e-4, weight decay 1e-4 |
| LR schedule | Cosine annealing to 0 | Cosine annealing to 0 |
| Gradient clipping | 1.0 | 1.0 |
| Epochs | 200 | 200 |
| Batch size | 2 | 2 |
| Random seeds | Single model | Seeds 0-9 (weight init + data order) |
| Hardware | Google Colab, NVIDIA T4 (16 GB) | Google Colab Pro, NVIDIA A100 (40 GB) |
| Wall-clock training time | ~2.6 h (~47 s/epoch) | ~60 min/model, ~10 h total |

*Table 2. Training configuration for the two UQ campaigns.*

## 6. Results

### *6.1 MC Dropout: Learned Rates and Calibrated Bands*

The concrete dropout model converged to a best relative L2 training loss of 0.0564 at epoch 192. The 25 learned dropout rates, all initialized at 0.10, differentiated into a structured pattern during training (Figure 3): the GALE cross-attention output layers consistently learned the highest rates (0.119-0.143 across all eight blocks), the geometry tokenizer settled at 0.123, while the self-attention and feed-forward layers learned lower rates (0.078-0.090). The separation began near epoch 25 and stabilized by epoch 150, coinciding with the flattening of the training loss. The model autonomously concluded that the pathway merging geometry context into the physical states carries the greatest uncertainty, a physically interpretable and entirely emergent finding that required no manual tuning.

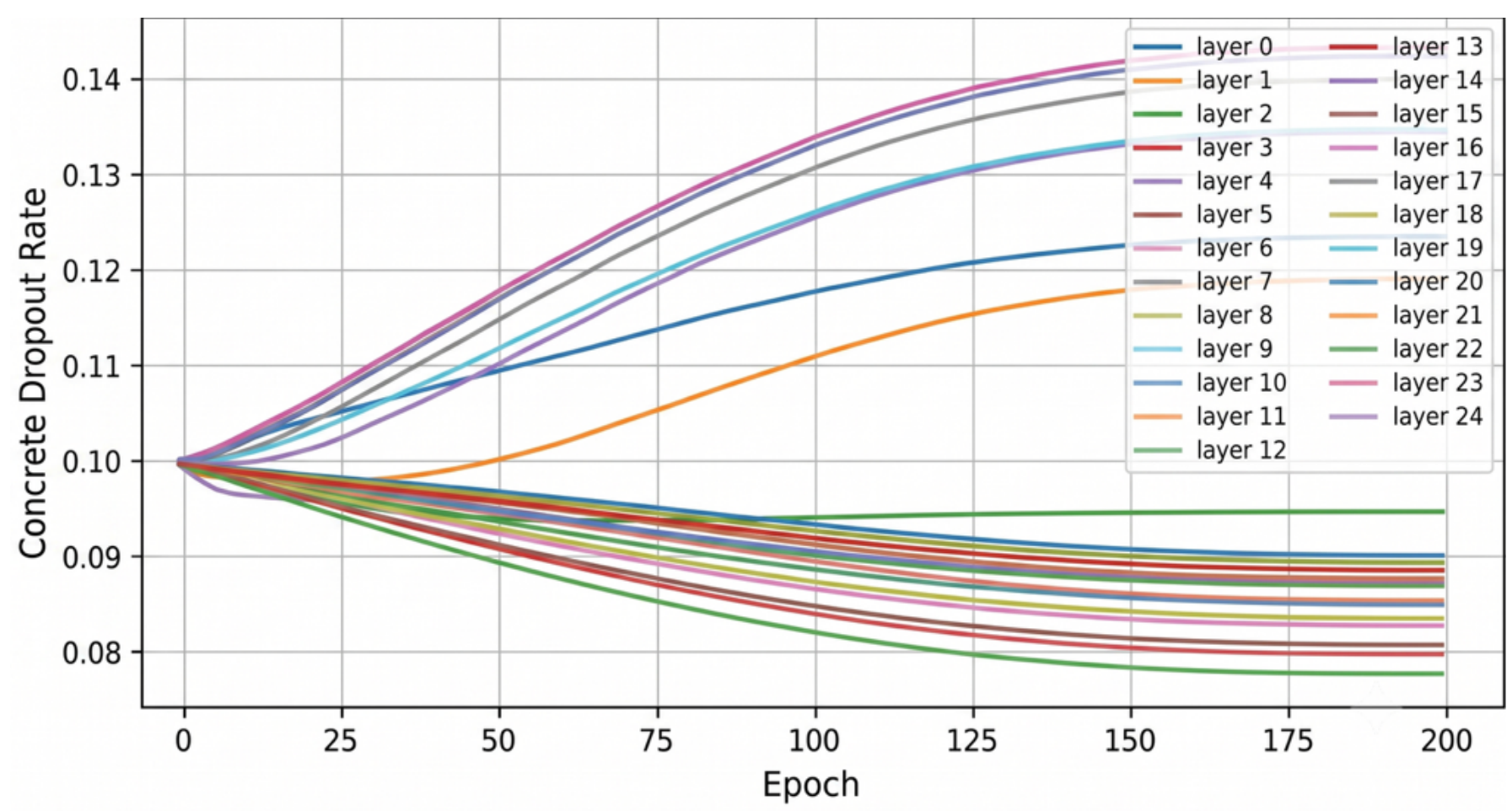

*Figure 3. Evolution of the 25 learned concrete dropout rates over 200 epochs. The upper cluster corresponds to GALE cross-attention output layers and the geometry tokenizer; the lower cluster to self-attention and feed-forward layers.*

On the five held-out simulations the deterministic (dropout-off) accuracy averaged 4.50% relative L2 error (range 3.35-5.50%). With 50 stochastic passes, the ±2σ band achieved 100% empirical coverage on every test run, with mean band widths of 12.6-15.3 mm (Table 3). The bands exhibit physically meaningful structure (Figure 4): they are tight during the well-constrained early loading phase, widen through the strongly nonlinear crush phase, and remain wide in the late-time regime (t = 60-100) where transient surrogates of this family are known to drift. Notably, the widest band (15.27 mm) coincides with the least accurately predicted run (Exp_83, 5.50% error), and the tightest band (12.56 mm) with one of the better-predicted runs (Exp_5, 4.86% error), indicating partial self-consistency between band width and prediction difficulty. The clearest correspondence is at the worst-case end: the model is most uncertain where it is least accurate, which is the behavior required for useful engineering deployment.

| Test Run | t_DP600 (mm) | t_DP1000 (mm) | Rel. L2 Error (%) | Coverage (%) | Mean Band Width (mm) |
|---|---|---|---|---|---|
| Exp_16 | 1.699 | 2.213 | 3.82 | 100 | 14.49 |
| Exp_33 | 2.773 | 2.225 | 4.99 | 100 | 14.12 |
| Exp_37 | 1.999 | 2.158 | 3.35 | 100 | 14.33 |
| Exp_5 | 2.788 | 2.187 | 4.86 | 100 | 12.56 |
| Exp_83 | 2.618 | 2.312 | 5.50 | 100 | 15.27 |
| Mean | - | - | 4.50 | 100.0 | 13.75 |

*Table 3. MC Dropout (concrete) results on the five held-out OpenRadioss simulations. Coverage is the percentage of timesteps at which the FE truth lies within the ±2σ band at node 1806.*

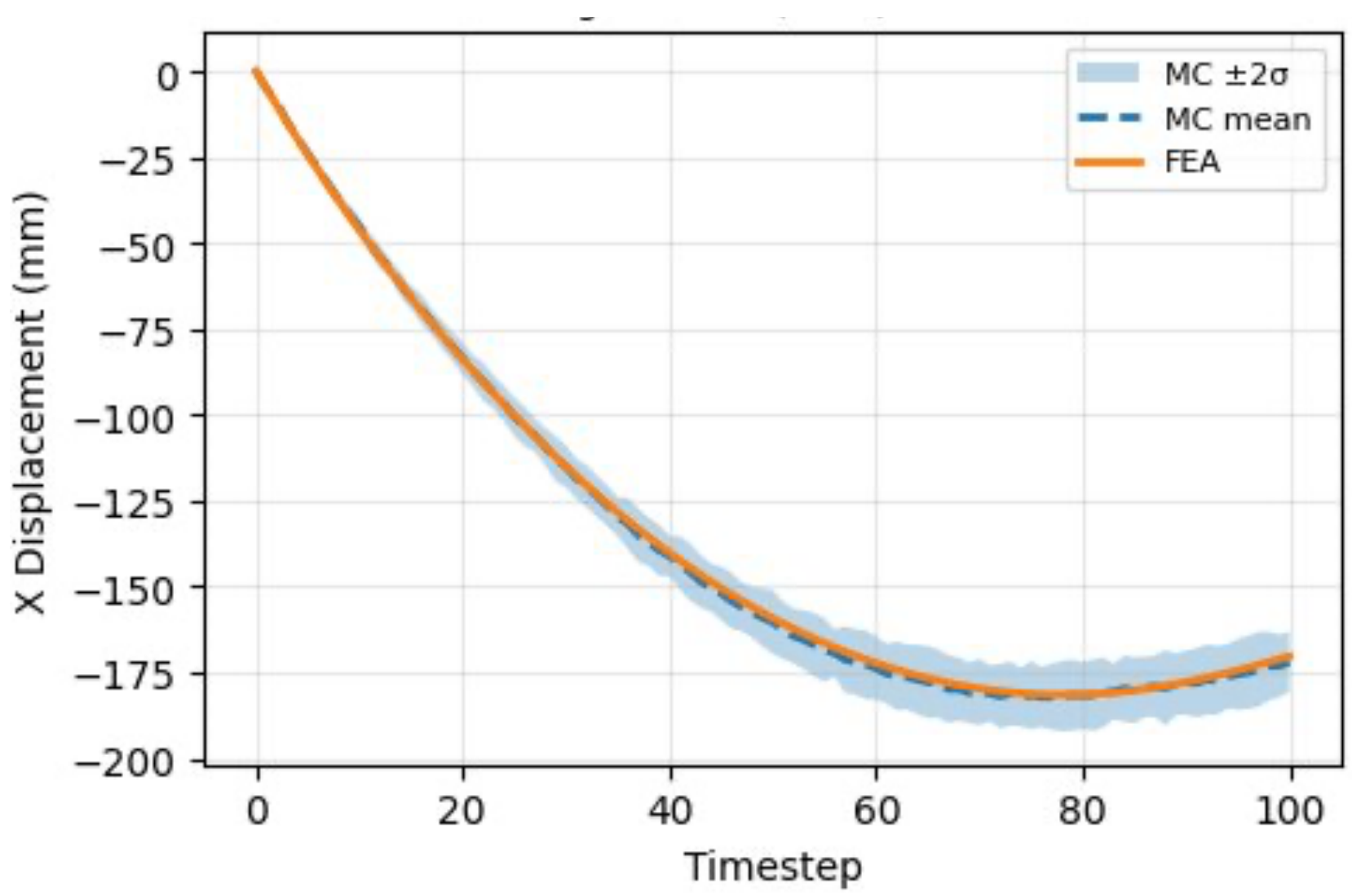

*Figure 4. MC Dropout uncertainty band at node 1806 (test run Exp_16, 50 stochastic passes). The FE ground truth (solid) remains inside the ±2σ band (shaded) over the full event; the band widens in the late-deformation regime.*

### *6.2 Deep Ensemble: Accurate but Overconfident*

Table 4 reports per-member statistics for the ten-model ensemble. Nine of the ten members converged tightly, with best training losses between 0.0222 and 0.0228 and held-out errors between 2.45% and 2.50%, roughly half the error of the dropout-regularized model. One member (seed 3) converged poorly, achieving its best loss of 0.0419 at epoch 35 and a test error of 6.69%. The outlier was deliberately retained: occasional convergence to inferior basins is an inherent property of the ensemble mechanism, and its disagreement with the well-converged members is precisely the signal the method is designed to capture.

| Seed | Best Training Loss | Best Epoch | Mean Test Error (%) |
|---|---|---|---|
| 0 | 0.02255 | 194 | 2.50 |
| 1 | 0.02254 | 195 | 2.47 |
| 2 | 0.02269 | 198 | 2.47 |
| 3 | 0.04192 | 35 | 6.69 |
| 4 | 0.02281 | 195 | 2.49 |
| 5 | 0.02264 | 197 | 2.45 |
| 6 | 0.02261 | 198 | 2.45 |
| 7 | 0.02258 | 199 | 2.46 |
| 8 | 0.02240 | 198 | 2.46 |
| 9 | 0.02222 | 199 | 2.47 |

*Table 4. Per-member training and test statistics for the ten-model deep ensemble (seeds 0-9).*

Despite the superior point accuracy (ensemble mean error 2.89%, or approximately 2.47% excluding the outlier member), the ensemble uncertainty bands are severely overconfident (Table 5, Figure 5). Empirical coverage ranged from 5% to 67% across the five test runs, averaging 42.2%, far below the 95% target. The mechanism is visible in the per-run pattern: on Exp_16 and Exp_37 the nine well-converged members agree almost perfectly with one another (band widths of only 2.4-3.7 mm) while sharing a common systematic bias relative to the FE truth, which therefore falls outside the narrow band at nearly every timestep. On Exp_33, Exp_5, and Exp_83 the outlier member inflates the band to 10-11 mm, partially rescuing coverage to 65-67%, but still well short of target.

| Test Run | Mean Error Across 10 Models (%) | Coverage (%) | Mean Band Width (mm) |
|---|---|---|---|
| Exp_16 | 2.37 ± 0.70 | 5 | 2.43 |
| Exp_33 | 3.06 ± 1.65 | 67 | 11.09 |
| Exp_37 | 2.50 ± 0.87 | 6 | 3.67 |
| Exp_5 | 2.97 ± 1.62 | 67 | 11.06 |
| Exp_83 | 3.56 ± 1.50 | 65 | 10.01 |
| Mean | 2.89 | 42.2 | 7.65 |

*Table 5. Deep ensemble results on the five held-out simulations.*

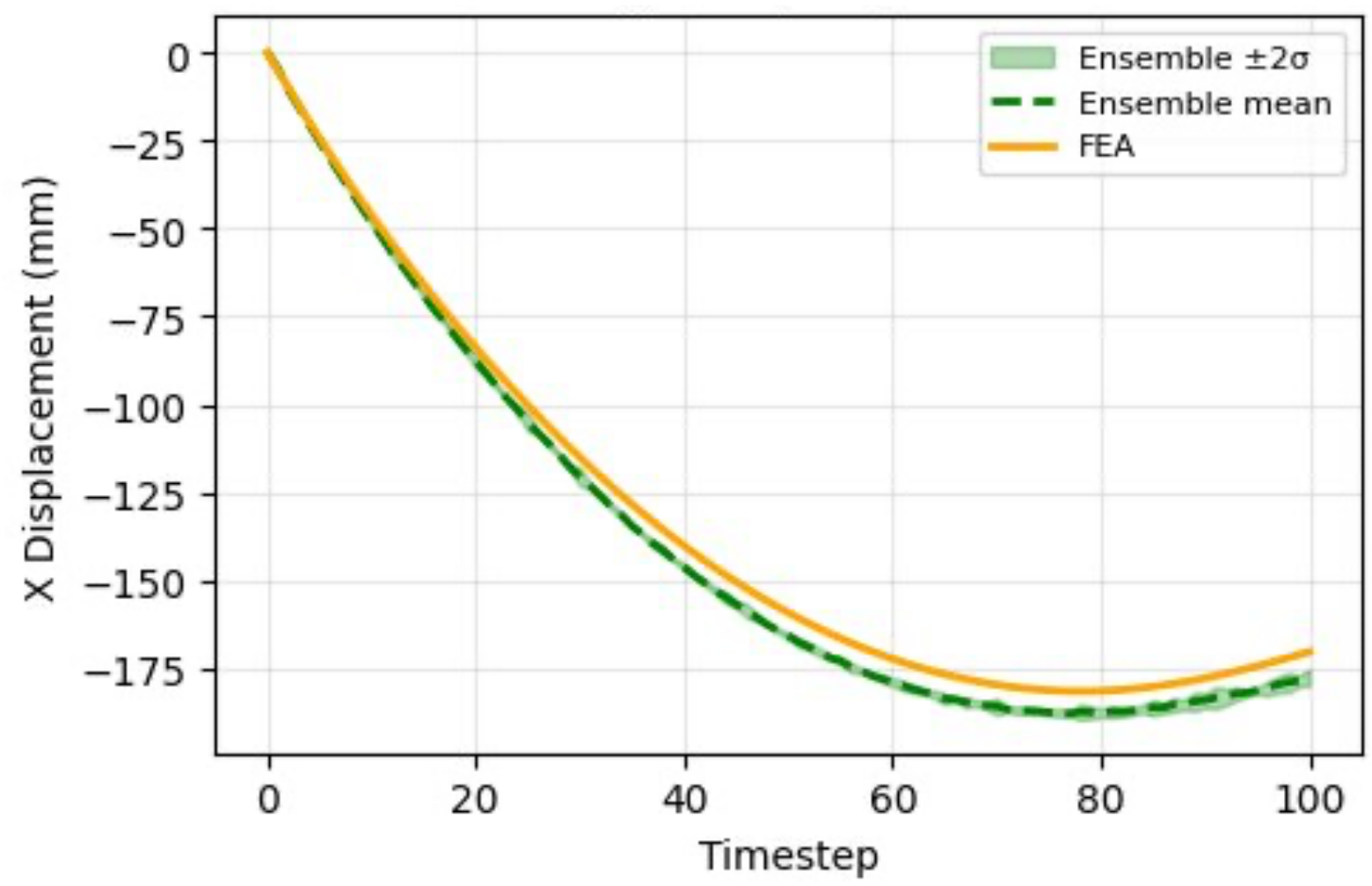

*Figure 5. Deep ensemble band at node 1806 (test run Exp_16). The ten members agree closely with one another but share a common bias; the FE truth (solid) escapes the narrow ±2σ band over most of the event (coverage 5%).*

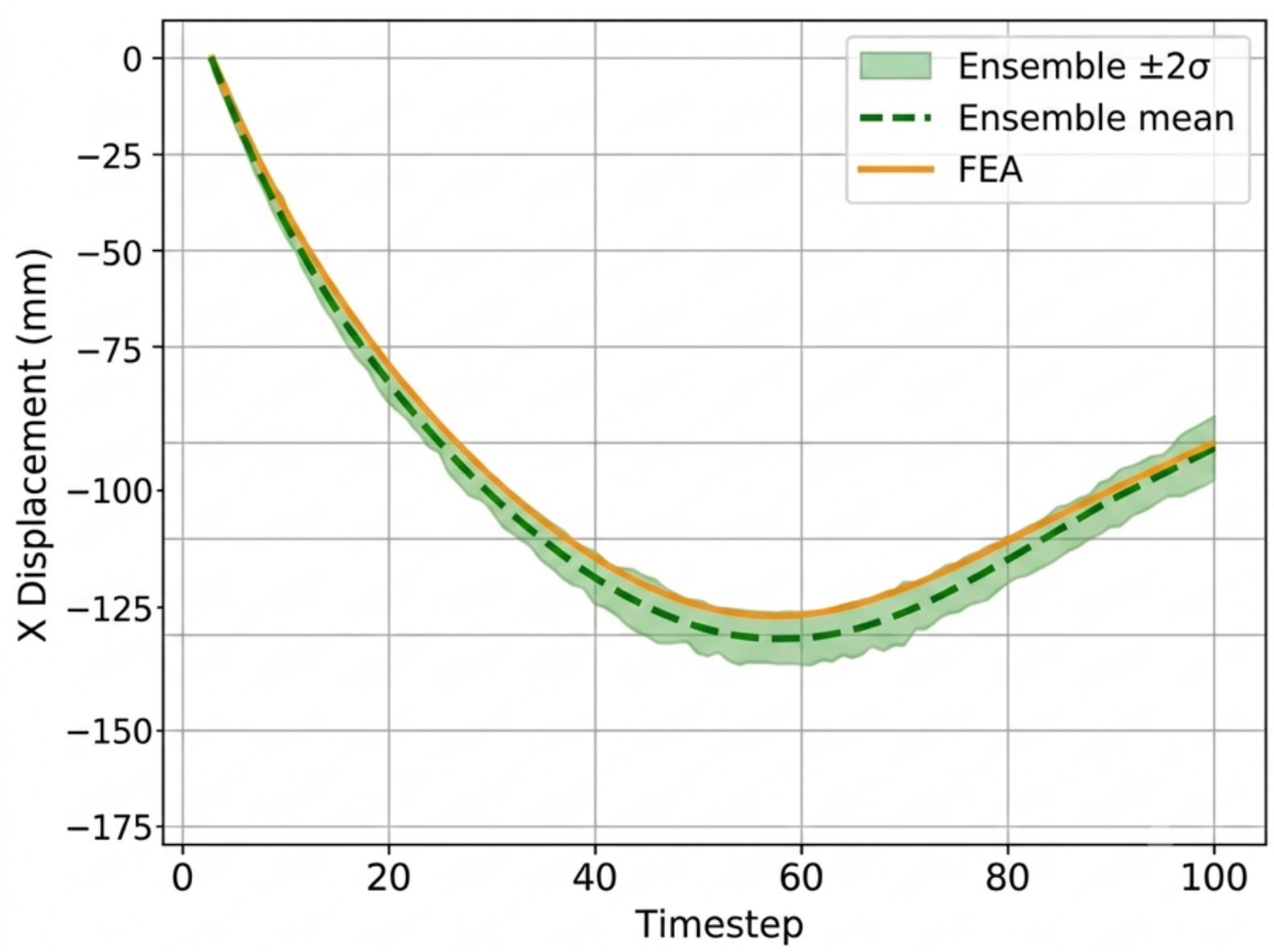

*Figure 6. Deep ensemble band on Exp_83. The FE truth exits the band in the late-deformation regime (t > 50), where all members drift in the same direction (coverage 65%).*

### *6.3 Head-to-Head Comparison*

Table 6 consolidates the comparison. The two methods occupy opposite corners of the accuracy-calibration trade space: the ensemble produces the better point estimate, while MC Dropout produces the only trustworthy uncertainty estimate, and does so at one-tenth the training cost.

| Metric | MC Dropout (Concrete) | Deep Ensemble (10 models) |
|---|---|---|
| Mean test error (%) | 4.50 | 2.89 |
| Empirical coverage at ±2σ (%) | 100.0 | 42.2 |
| Mean band width (mm) | 13.75 | 7.65 |
| Training runs required | 1 | 10 |
| Inference passes | 50 (stochastic) | 10 (deterministic) |
| GPU training cost | ~2.6 T4-hours | ~10 A100-hours |
| Dropout-rate selection | Learned (concrete relaxation) | Not applicable |
| Calibration verdict | Conservative, well-calibrated | Overconfident (under-covers) |

*Table 6. Head-to-head comparison of the two UQ methods on the identical architecture, dataset, and test split.*

## 7. Discussion

The central finding, that a ten-member deep ensemble underestimates uncertainty by more than a factor of two in coverage terms while MC Dropout with learned rates achieves full coverage, runs counter to the prevailing assumption that deep ensembles are the universal gold standard for deep learning UQ [13]. The explanation lies in what each method's variance actually measures. Ensemble variance captures disagreement between independently trained solutions; when a smooth, low-dimensional design space (here, two thickness variables) and a sufficiently expressive architecture lead nearly all members into closely equivalent basins, the members agree confidently, even where they are collectively wrong. Ensemble spread cannot, by construction, detect a shared systematic bias. The observed failure mode on Exp_16 and Exp_37, nine members in near-perfect mutual agreement with a 2-4 mm band that misses the FE truth at 94-95% of timesteps, is exactly this mechanism.

MC Dropout's stochasticity, by contrast, perturbs the internal representation of a single converged model. Because the concrete dropout rates were learned in the same optimization that fit the data, the perturbation scale reflects how much internal redundancy the model could afford at each layer while still fitting the training set, and the resulting band inflates wherever the prediction relies on weakly constrained pathways. The emergent structure of the learned rates, with the geometry-fusion (GALE cross-attention) layers carrying the highest rates, suggests that the dominant uncertainty in this surrogate concerns how geometric context modulates the temporal response, which is consistent with the late-time band widening observed on every test run.

Several practical implications follow for engineering deployment. First, calibration must be measured, not assumed: a method's reputation is not a substitute for empirical coverage on held-out simulations. Second, for surrogate problems characterized by smooth design spaces and strongly converging training runs, ensemble bands should be treated as a lower bound on uncertainty and supplemented or recalibrated before use in design decisions. Third, the cost asymmetry matters in practice: the MC Dropout study consumed roughly 2.6 GPU-hours on freely available hardware, whereas the ensemble required ten full training runs on a premium accelerator. For practitioners operating under the zero-cost constraint that motivates this work, the better-calibrated method is also the cheaper one. Fourth, conservative bands have direct engineering value: a 100%-coverage band of approximately 14 mm on a 180 mm peak intrusion (roughly ±4% of the signal at the $\pm 2\sigma$ level) is tight enough to support screening decisions while guaranteeing, on this test set, that the FE truth is never outside it.

Limitations of the present study suggest future work. The benchmark involves a single component with a two-variable parametric design space; richer geometric variation may increase ensemble diversity and partially close the calibration gap. The coverage statistics derive from five held-out runs at a single KPI node; full-field coverage maps and larger test sets would strengthen the statistical claims. Hybrid approaches (ensembles of dropout models), post-hoc recalibration, and conformal prediction methods that provide distribution-free coverage guarantees are natural extensions. Finally, the PhysicsNeMo Guardrails module, which estimates confidence from checkpoint ensembles and training-distribution similarity, offers a complementary system-level perspective that the author intends to evaluate in subsequent work.

## 8. Conclusions

This work compared Monte Carlo Dropout with learned concrete rates against a ten-member deep ensemble for uncertainty quantification of a transient crash surrogate, using an identical GeoTransolver architecture, an identical OpenRadioss bumper beam dataset, and an entirely open-source, zero-cost toolchain. Three conclusions emerge:

1. MC Dropout with concrete (learned) rates produced well-calibrated, conservative uncertainty bands, achieving 100% empirical coverage of the FE ground truth on all five held-out simulations with a mean band width of 13.75 mm, at the cost of a single training run. The learned per-layer rates eliminated the dropout-rate tuning problem that historically limited the method, and their emergent structure localized the dominant model uncertainty to the geometry-fusion pathways of the architecture.

2. The deep ensemble delivered superior point accuracy (2.89% vs. 4.50% mean relative L2 error) but severely overconfident uncertainty, covering the FE truth at only 42.2% of timesteps on average, because independently trained members converged to closely equivalent solutions sharing a common systematic bias that ensemble spread cannot detect.

3. For surrogate modeling problems of this character, smooth low-dimensional design spaces with strongly converging training, the assumption that deep ensembles constitute the default gold standard for UQ deserves scrutiny. Calibration should be verified empirically, and learned-rate MC Dropout merits consideration as the primary UQ mechanism, particularly under compute constraints.

The complete pipeline, an open-source explicit FE solver, an open-source physics-ML framework, and free-tier cloud GPUs, demonstrates that rigorous uncertainty quantification for crash surrogates is accessible without commercial software, lowering the barrier for both industrial evaluation and academic instruction.

## References

[1] Altair Engineering, "OpenRadioss: Open-Source Explicit Finite Element Solver," https://openradioss.org, accessed 2026.

[2] NVIDIA Corporation, "PhysicsNeMo: An Open-Source Framework for Physics-Based Machine Learning," https://github.com/NVIDIA/physicsnemo, 2026.

[3] Nabian, M.A., Chavare, S., Akhare, D., Ranade, R., Cherukuri, R., and Tadepalli, S., "Automotive Crash Dynamics Modeling Accelerated with Machine Learning," arXiv preprint arXiv:2510.15201, 2025.

[4] Akhare, D., Nabian, M.A., Adams, C., Chavare, S., and Choudhry, S., "High-Fidelity Industrial Crash Dynamics Prediction via Geometry-Aware Operator Learning with Memory-Efficient Low-Rank Attention," arXiv preprint arXiv:2605.27758, 2026.

[5] Wu, H., Luo, H., Wang, H., Wang, J., and Long, M., "Transolver: A Fast Transformer Solver for PDEs on General Geometries," Proceedings of the 41st International Conference on Machine Learning (ICML), 2024. arXiv:2402.02366.

[6] Adams, C., Ranade, R., Cherukuri, R., and Choudhry, S., "GeoTransolver: Learning Physics on Irregular Domains Using Multi-scale Geometry Aware Physics Attention Transformer," arXiv preprint arXiv:2512.20399, 2025.

[7] Srivastava, N., Hinton, G., Krizhevsky, A., Sutskever, I., and Salakhutdinov, R., "Dropout: A Simple Way to Prevent Neural Networks from Overfitting," Journal of Machine Learning Research 15(1):1929-1958, 2014.

[8] Gal, Y. and Ghahramani, Z., "Dropout as a Bayesian Approximation: Representing Model Uncertainty in Deep Learning," Proceedings of the 33rd International Conference on Machine Learning (ICML), 2016.

[9] Gal, Y., Hron, J., and Kendall, A., "Concrete Dropout," Advances in Neural Information Processing Systems 30 (NeurIPS), 2017.

[10] Kendall, A. and Gal, Y., "What Uncertainties Do We Need in Bayesian Deep Learning for Computer Vision?" Advances in Neural Information Processing Systems 30 (NeurIPS), 2017.

[11] Hansen, L.K. and Salamon, P., "Neural Network Ensembles," IEEE Transactions on Pattern Analysis and Machine Intelligence 12(10):993-1001, 1990.

[12] Breiman, L., "Bagging Predictors," Machine Learning 24(2):123-140, 1996.

[13] Lakshminarayanan, B., Pritzel, A., and Blundell, C., "Simple and Scalable Predictive Uncertainty Estimation Using Deep Ensembles," Advances in Neural Information Processing Systems 30 (NeurIPS), 2017.

[14] Fort, S., Hu, H., and Lakshminarayanan, B., "Deep Ensembles: A Loss Landscape Perspective," arXiv preprint arXiv:1912.02757, 2019.

[15] Guo, C., Pleiss, G., Sun, Y., and Weinberger, K.Q., "On Calibration of Modern Neural Networks," Proceedings of the 34th International Conference on Machine Learning (ICML), 2017.

[16] Chavare, S. and Mourelatos, Z., "Uncertainty Quantification in Machine Learning Using an Ensemble Approach with Gaussian Process Regression," WCX SAE World Congress Experience, Detroit, Michigan, United States, April 8, 2025, https://doi.org/10.4271/2025-01-8199.

## Acknowledgments

This work was conducted in the author's personal capacity, and is independent of any employer. The author acknowledges the OpenRadioss community for the open-source bumper beam model and the NVIDIA PhysicsNeMo team for the open-source GeoTransolver implementation and concrete dropout module. Anthropic's Claude was used to assist with code development, analysis workflow, and manuscript drafting; all results were generated, verified, and interpreted by the author.

## Definitions / Abbreviations

| Term | Definition |
| --- | --- |
| CAE | Computer-Aided Engineering |
| DoE | Design of Experiments |
| FE / FEA | Finite Element / Finite Element Analysis |
| GALE | Geometry-Aware Local Embeddings (GeoTransolver cross-attention) |
| KPI | Key Performance Indicator |
| MC Dropout | Monte Carlo Dropout |
| ML | Machine Learning |
| UQ | Uncertainty Quantification |
| VTP | VTK PolyData file format |